\RequirePackage{iftex}
\ifPDFTeX
\RequirePackage{cmap}
\fi
\documentclass[14pt,a4paper]{extarticle}

\ifPDFTeX
  \usepackage[T2A,T1]{fontenc}
  \usepackage[utf8]{inputenc}
\else
  \usepackage{fontspec}
\fi
\usepackage[english,russian]{babel}
\usepackage[a4paper,top=2.5cm,bottom=2.5cm,left=2.5cm,right=2.5cm]{geometry}
\usepackage{setspace}
\usepackage{indentfirst}
\usepackage{amsmath,amssymb}
\usepackage{array}
\usepackage{makecell}
\usepackage{enumitem}
\usepackage{graphicx}
\usepackage{caption}
\usepackage{float}
\usepackage{textcomp}
\usepackage{xcolor}
\usepackage{hyperref}
\usepackage{tikz}
\usetikzlibrary{shapes.geometric,arrows.meta,positioning,calc,fit,backgrounds}

\hypersetup{
  colorlinks=true,
  linkcolor=black,
  citecolor=black,
  urlcolor=black
}
\addto\captionsenglish{}

\newcolumntype{P}[1]{>{\raggedright\arraybackslash\hspace{0pt}}p{#1}}
\newcolumntype{C}[1]{>{\centering\arraybackslash\hspace{0pt}}p{#1}}
\newcommand\thinhline{\Xhline{0.1pt}}

\begin{document}
\selectlanguage{english}

{\noindent\textbf{UDC 616.5-006:004.85}}

\bigskip
\noindent\textbf{Corresponding author:} Elena S. Kozachok, e.kozachok@ispras.ru, +7~906~571~0071.

\bigskip
\begin{center}
\textbf{\textit{Clinical Validation of the Melanoscope~AI Mobile Dermoscopy Clinical Decision Support System}}
\end{center}

\bigskip
\noindent Elena S.~Kozachok$^{1}$, Sergey S.~Seregin$^{2}$

\medskip
\noindent$^{1}$Ivannikov Institute for System Programming of the Russian Academy of Sciences (ISP~RAS),\\
25 Alexander Solzhenitsyn str., Moscow, 109004, Russian Federation\\
$^{2}$Orel Regional Oncology Dispensary,\\
2 Ippodromny lane, Orel, Russian Federation

\bigskip
\noindent{\itshape Abstract.

\textbf{Introduction/Background.} Early detection of malignant skin lesions is critical for prognosis, yet dermatologist shortages in Russian regions limit screening coverage. Mobile dermoscopy clinical decision support systems (CDSS) offer a~promising approach, with model interpretability and standardised patient routing remaining key barriers to adoption.

\textbf{Aim.} To develop a~quantitative interpretability assessment method for cascade deep learning models and a~three-zone patient routing algorithm, and to conduct a~preliminary single-centre prospective clinical validation with independent expert assessment of the \emph{Melanoscope~AI} CDSS in Russian outpatient practice.

\textbf{Material and methods.} A~two-stage cascade classification of dermoscopic images; attention map visualisation (attention rollout for ViT and Swin; Grad-CAM for ConvNeXt and EfficientNetV2); quantitative assessment of agreement between activation maps and expert dermoscopic structure annotations via the intersection-over-union (IoU) metric; a~prospective single-centre observational clinical validation in four ``Melanoma Day'' preventive screening sessions (Beauty Clinic, Orel, Russia, June~2025\,--\,April~2026); comparison of automatic classification with an~independent expert assessment and histological verification of all detected malignant lesions.

\textbf{Results.} On a~sample of 176~patients and 176~primary-lesion dermoscopic images the agreement between the automatic classification and the expert reached 88.6\,\%; no false negatives were observed among 5 malignant lesions (5 of 5; 95\,\% confidence interval (CI): 47.8\,--\,100.0\,\%), specificity\,---\,88.3\,\%. Three melanomas and two basal cell carcinomas were detected and histologically confirmed; six dysplastic naevi were placed under dynamic follow-up. Mean IoU values on the annotated subset ($n=180$): ViT\,---\,0.69; Swin\,---\,0.64; ConvNeXt\,---\,0.53; EfficientNetV2\,---\,0.51. A~three-zone patient routing algorithm with thresholds $P<0.15$\,/\,$0.15$\,--\,$0.50$\,/\,$\geq 0.50$ was developed.

\textbf{Conclusion.} No false negatives were observed among 5 malignant lesions in this preliminary single-centre validation (95\,\% CI: 47.8\,--\,100.0\,\%); specificity was 88.3\,\%, supporting use in a~screening setting. The integrated chain of cascade classification, attention map visualisation with IoU assessment, and a~three-zone routing algorithm provides reproducible interpretable clinical decision support adaptable to facilities with varying resource levels.}


\medskip
\noindent\textbf{Funding:} the study was conducted without financial support from grants, public, non-profit, or commercial organisations and entities.

\medskip
\noindent\textbf{Resources:} computational tasks were performed using the cloud platform resources of ISP~RAS.

\medskip
\noindent\textbf{Compliance with ethical standards:} the study was conducted in accordance with the Declaration of Helsinki. All participants provided informed consent for the processing of anonymised clinical data and dermoscopic images. The study was observational in design and did not alter the standard patient care pathway.

\bigskip
\section*{Introduction}\label{sec:intro}

Melanoma accounts for less than 5\,\% of all malignant skin lesions but contributes approximately 80\,\% of deaths in this group; five-year survival when detected at stage~I reaches 98\,\%, versus approximately 23\,\% at stage~IV~\cite{sung2021globocan,seer2023}. Early visual diagnosis provides a~critical time advantage before treatment initiation; however, the accuracy of melanoma recognition by a~general practitioner (GP) without dermoscopy does not exceed 70\,--\,75\,\%, whereas a~trained dermatologist with a~dermatoscope achieves 85\,--\,90\,\%~\cite{kittler2002}. In the Russian primary-care network, where the density of dermatologists in regions is substantially lower than in major cities, the task of instrumental support for the GP's diagnostic decision acquires direct practical significance.

Deep learning--based clinical decision support systems (CDSS) are capable of achieving diagnostic accuracy at the level of a~dermatologist~\cite{esteva2017,brinker2019,tschandl2020hcc}. Commercial solutions\,---\,SkinVision, Google Dermatology Assist, Botkin.AI, ProRodinka\,---\,have been evaluated in various clinical settings~\cite{udrea2020,liu2020}. The published literature, however, identifies three persistent limitations that impede the adoption of such systems in Russian clinical practice.

\textbf{First limitation\,---\,opacity of the automatic decision.} Most commercial CDSSs return only a~probabilistic prediction or categorical diagnosis to the clinician, without visualising the features that drove the classification. This prevents expert oversight and fosters clinician distrust in the system as a~clinical decision tool~\cite{tonekaboni2019}. In the context of medical AI, interpretability is regarded not as an optional feature but as a~mandatory prerequisite for safe clinical use~\cite{reyes2020}.

\textbf{Second limitation\,---\,absence of standardised routing.} The result of automatic classification is usually presented either as a~continuous probability or a~categorical class, but does not specify a~concrete clinical action depending on the risk level. Without a~formalised routing scheme, a~CDSS cannot be integrated into the workflow of a~resource-limited institution, since the staff action algorithm remains at the discretion of each individual specialist and loses reproducibility.

\textbf{Third limitation\,---\,absence of independent validations in the Russian population.} The majority of validation publications on AI-based CDSSs in dermoscopy have been conducted in Western populations with Fitzpatrick skin phototypes II\,--\,III~\cite{combalia2022}; results for phototypes I\,--\,IV typical of the Russian population are virtually absent from the published literature.

\textbf{Aim of the study}\,---\,to develop a~quantitative interpretability assessment method for cascade CDSS models and a~three-zone patient routing algorithm; to conduct a~prospective clinical validation with independent expert assessment of the new edition of the \emph{Melanoscope AI} CDSS~\cite{kozachok2025sppvr1,kozachok2025sppvr2}; and to formulate recommendations for integrating the system into medical practice.

\textbf{Scientific novelty} of the study:
\begin{enumerate}
\item A~new composition of the \emph{Melanoscope AI} CDSS is described, comprising a~mobile image-acquisition application, a~server-side cascade inference subsystem, and an attention-map visualisation module with quantitative assessment of clinical relevance.
\item A~quantitative method for assessing the interpretability of cascade models is developed, based on the IoU metric between the model's high-activation map and expert annotations of dermoscopic structures; the method is applied to four architectures (Vision Transformer (ViT), Swin Transformer (Swin), ConvNeXt, EfficientNetV2).
\item A~prospective clinical validation with independent expert assessment was conducted in Russian outpatient practice (4 sessions, 2025\,--\,2026, Orel) with histological verification of all malignant lesions.
\item A~three-zone patient routing algorithm with clinically justified malignancy probability thresholds $P<0.15$\,/\,$0.15$\,--\,$0.50$\,/\,$\geq 0.50$ was developed.
\item Integration recommendations were formulated for three types of healthcare facilities with varying resource levels.
\end{enumerate}

\textbf{Relationship to prior publications.} The methodology for creating a~clinically verified dermoscopic image dataset with multi-stage expert verification is described in~\cite{kozachok2026dataset}; the preceding dataset with annotation of clinically significant features\,---\,in~\cite{kozachok2025dataset}. The screening examination methodology using mobile dermoscopy\,---\,in~\cite{kozachok2025screening}. The architectural description of the intelligent CDSS for skin lesion diagnosis based on dermoscopic image analysis\,---\,in~\cite{kozachok2025sppvr1}. A~brief characterisation of the CDSS with mobile dermoscopy\,---\,in~\cite{kozachok2025sppvr2}. The CDSS mobile application is registered as an independent computer program~\cite{kozachok2026mobile}. Differences between the present work and the listed publications are discussed in detail in Section~\ref{sec:diff}.

\section{Materials and Methods}\label{sec:methods}

\subsection{The \emph{Melanoscope~AI} Clinical Decision Support System}\label{sec:sppvr}

\subsubsection{System Architecture}\label{sec:components}

\begin{figure}
\centering
\includegraphics[width=\linewidth]{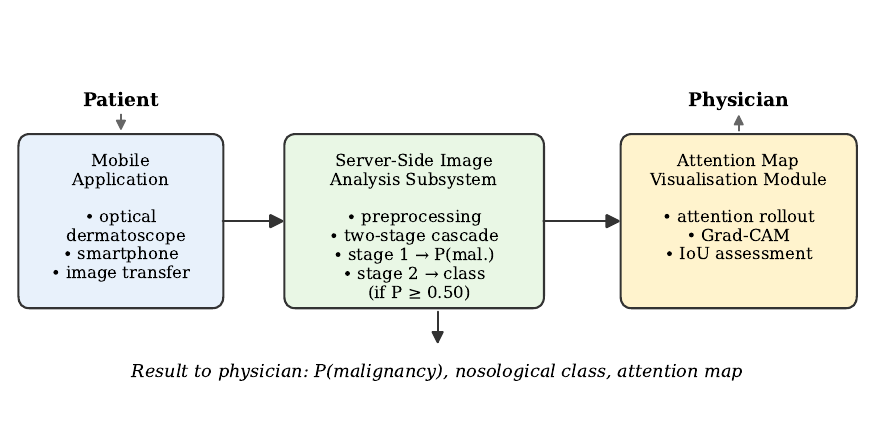}
\caption{Architecture of the \emph{Melanoscope~AI} CDSS: mobile image-acquisition application, server-side cascade inference subsystem and attention-map visualisation module. Arrows show data flows during a~standard examination cycle.}\label{fig:arch}
\end{figure}

The \emph{Melanoscope AI} CDSS was developed as a~computer program in accordance with the requirements of GOST~R~71671-2024 on clinical decision support systems employing artificial intelligence~\cite{kozachok2026reg}. The system implements a~two-stage cascade classification of dermoscopic images and comprises three functional components shown in Figure~\ref{fig:arch}:

\begin{enumerate}
\item \textbf{Mobile application}~\cite{kozachok2026mobile} for acquiring dermoscopic images using an optical dermatoscope coupled to a~smartphone, registering an anonymised patient record, and transmitting the image to the server.
\item \textbf{Server-side image analysis subsystem}, performing preprocessing, cascade-model inference, and returning the probabilistic prediction together with the attention map.
\item \textbf{Attention-map visualisation module}, generating an activation heatmap for each classification result and assessing its agreement with clinically relevant image regions.
\end{enumerate}

Unlike stationary dermoscopic workstations and web-based solutions, the mobile image-acquisition format makes the system applicable at screening events and outpatient offices without specialised stationary equipment.

\subsubsection{Cascade Classification}\label{sec:cascade}

Classification of a~dermoscopic image is organised in two stages:
\begin{itemize}
\item \textbf{Stage~1}\,---\,binary separation of malignant and benign lesions; output\,---\,malignancy probability~$P$.
\item \textbf{Stage~2}\,---\,differentiation of the nosological class when $P\geq 0.50$: melanoma (MEL), squamous cell carcinoma (SCC), basal cell carcinoma (BCC).
\end{itemize}

Comparative analysis of architectures for both stages, domain-adaptation strategies, and training metrics are the subject of a~separate study and are not reported here.

\subsubsection{Clinical Interface}\label{sec:interface}

The clinician receives three result components: the malignancy probability $P$ at the output of cascade Stage~1; the refined nosological class at the output of Stage~2 (when triggered); and an attention heatmap highlighting the image regions that determined the model's decision. Together these components provide not only the prediction but also a~formalised explanation of what in the image served as the basis for the classification.

\subsubsection{Differences from Prior Publications}\label{sec:diff}

It should be noted that previously published works~\cite{kozachok2025sppvr1,kozachok2025sppvr2} describe a~different software system whose rights belong to Open Innovations LLC; the \emph{Melanoscope AI} system presented in this article is an independent new development of ISP~RAS~\cite{kozachok2026reg}. Unlike prior publications~\cite{kozachok2025dataset,kozachok2025screening,kozachok2025sppvr1,kozachok2025sppvr2}, the present work for the first time presents a~clinically validated edition of the system with verified diagnostic accuracy, quantitative IoU-based interpretability assessment of attention maps, a~formalised three-zone routing algorithm, and histological verification of all detected malignant lesions.

\subsection{Quantitative Interpretability Assessment Method}\label{sec:interp}

\subsubsection{Attention Map Visualisation}

For each of the four architectures studied, a~visualisation method appropriate to its architectural type is applied.

\textbf{Attention rollout} (ViT-B/16, Swin-T)~\cite{abnar2020}. The attention map is formed by sequentially multiplying the attention matrices across all transformer blocks, averaging over heads and accounting for residual connections:
\begin{equation}
A_{\text{roll}} = \prod_{l=1}^{L} \left( 0.5\cdot\bar{A}_l + 0.5\cdot I \right),
\label{eq:rollout}
\end{equation}
where $\bar{A}_l$ is the head-averaged attention matrix of block~$l$, $I$ is the identity matrix, and $\prod$ denotes sequential left-to-right matrix multiplication from block $l=1$ to block $l=L$ (i.e.\ $M_L\cdot\ldots\cdot M_2\cdot M_1$, where $M_l=0.5\cdot\bar{A}_l+0.5\cdot I$). The resulting map $A_{\text{roll}}$ contains the attention weights from the $[\text{CLS}]$ token to all patch tokens; it is interpolated to the input image size and normalised to $[0,1]$.

\textbf{Grad-CAM} (ConvNeXt-B, EfficientNetV2)~\cite{selvaraju2017}. The activation map is formed from the gradients of the predicted class~$c$ with respect to the activations of the final convolutional layer $A^k$:
\begin{equation}
L^{c}_{\text{Grad-CAM}} = \text{ReLU}\!\left(\sum_k \alpha^c_k\,A^k\right), \quad \alpha^c_k = \frac{1}{Z}\sum_{i,j}\frac{\partial y^c}{\partial A^k_{ij}},
\label{eq:gradcam}
\end{equation}
where $Z$ is a~normalisation constant. The map is interpolated to the input image size and normalised to $[0,1]$.

\subsubsection{Expert Annotations of Dermoscopic Structures}

In dataset~\cite{kozachok2026dataset} the field \texttt{dermoscopic\_structures} contains expert annotation of clinically significant dermoscopic patterns for each image: reticular network (typical and atypical), globules, pseudopods, blue--white veil, vascular structures (dotted, arborising, comma-shaped), and pigmented pseudopodial pattern. Annotations were performed by a~specialist dermatologist in the Colba annotation environment (ISP~RAS) and stored as a~set of bounding rectangles specifying the location of each structure in image coordinates.

\subsubsection{IoU Consistency Metric}

To quantify the correspondence between regions of high model activation and clinically relevant areas, the intersection-over-union (IoU) metric is introduced between the binary high-activation mask $M_{\text{model}}$ and the binary expert annotation mask $M_{\text{expert}}$:
\begin{equation}
\text{IoU} = \frac{|M_{\text{model}}\cap M_{\text{expert}}|}{|M_{\text{model}}\cup M_{\text{expert}}|}.
\label{eq:iou}
\end{equation}
The mask $M_{\text{model}}$ is constructed by thresholding: pixels are included whose normalised attention-map value exceeds threshold $\tau=0.5$. The mask $M_{\text{expert}}$ is built as the union of all annotated \texttt{dermoscopic\_structures} bounding boxes for the given image. The following clinical interpretation of metric values is adopted: $\text{IoU}>0.5$\,---\,the model focuses predominantly on clinically significant regions; $0.3\leq\text{IoU}\leq 0.5$\,---\,partial correspondence requiring visual review; $\text{IoU}<0.3$\,---\,the decision is driven by clinically irrelevant features, serving as a~flag for manual review.

\subsubsection{Interpretability Assessment Results}

IoU was computed on a~subset ($n=180$) of 1026 images from dataset~\cite{kozachok2026dataset} for which expert annotation of the \texttt{dermoscopic\_structures} field was available: 18 melanoma (MEL), 15 basal cell carcinoma (BCC), 16 dysplastic naevus (DN), and 131 typical naevus (NV) images. The resulting IoU values by architecture and nosological class are presented in Table~\ref{tab:iou} (mean $\pm$ standard deviation).

\begin{table}\footnotesize
\caption{IoU values by architecture and nosological class (mean$\,\pm\,$SD)}\label{tab:iou}
\centering
\setlength{\tabcolsep}{2pt}
\renewcommand{\arraystretch}{1.2}
\begin{tabular}{|P{42mm}|C{20mm}|C{20mm}|C{22mm}|C{18mm}|C{22mm}|}
\hline
Architecture & Melanoma & BCC & Dyspl. naevus & Naevus & Mean\\
\hline
ViT-B/16 (rollout) & 0.74$\,\pm\,$0.09 & 0.71$\,\pm\,$0.11 & 0.68$\,\pm\,$0.12 & 0.61$\,\pm\,$0.14 & \textbf{0.69$\,\pm\,$0.12}\\
\thinhline
Swin-T (rollout) & 0.69$\,\pm\,$0.10 & 0.67$\,\pm\,$0.13 & 0.63$\,\pm\,$0.11 & 0.58$\,\pm\,$0.15 & 0.64$\,\pm\,$0.13\\
\thinhline
ConvNeXt-B (Grad-CAM) & 0.58$\,\pm\,$0.13 & 0.56$\,\pm\,$0.14 & 0.52$\,\pm\,$0.13 & 0.47$\,\pm\,$0.16 & 0.53$\,\pm\,$0.14\\
\thinhline
EfficientNetV2 (Grad-CAM) & 0.55$\,\pm\,$0.12 & 0.54$\,\pm\,$0.15 & 0.50$\,\pm\,$0.14 & 0.45$\,\pm\,$0.17 & 0.51$\,\pm\,$0.15\\
\hline
\end{tabular}
\end{table}

The ViT-B/16 architecture shows the highest agreement with expert patterns across all nosological classes. The advantage of transformer architectures is explained by the global attention mechanism: unlike convolutional networks with a~local receptive field, ViT captures spatial relationships between distant image regions, which aligns with dermoscopic clinical practice where diagnostically significant patterns (atypical network, blue--white veil) may be located not only at the centre of the lesion. The highest IoU values were recorded for melanoma and BCC\,---\,the nosologies with the most pronounced specific patterns; for naevi with regular structure IoU is predictably lower, since the absence of a~distinct local pattern reduces the possibility of accurately aligning the activation map with the expert mask.

\begin{figure}
\centering
\includegraphics[width=\linewidth]{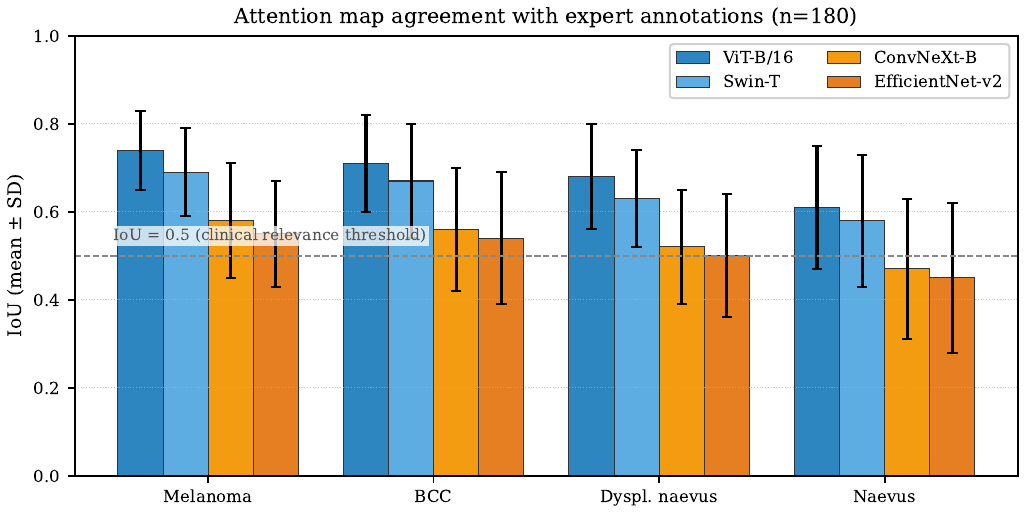}
\caption{Mean IoU values for four architectures and four nosological classes.}\label{fig:iou}
\end{figure}

A~graphical comparison of mean IoU values by architecture and nosological class is shown in Figure~\ref{fig:iou}.

\begin{figure}[t]
\centering
\includegraphics[width=\linewidth]{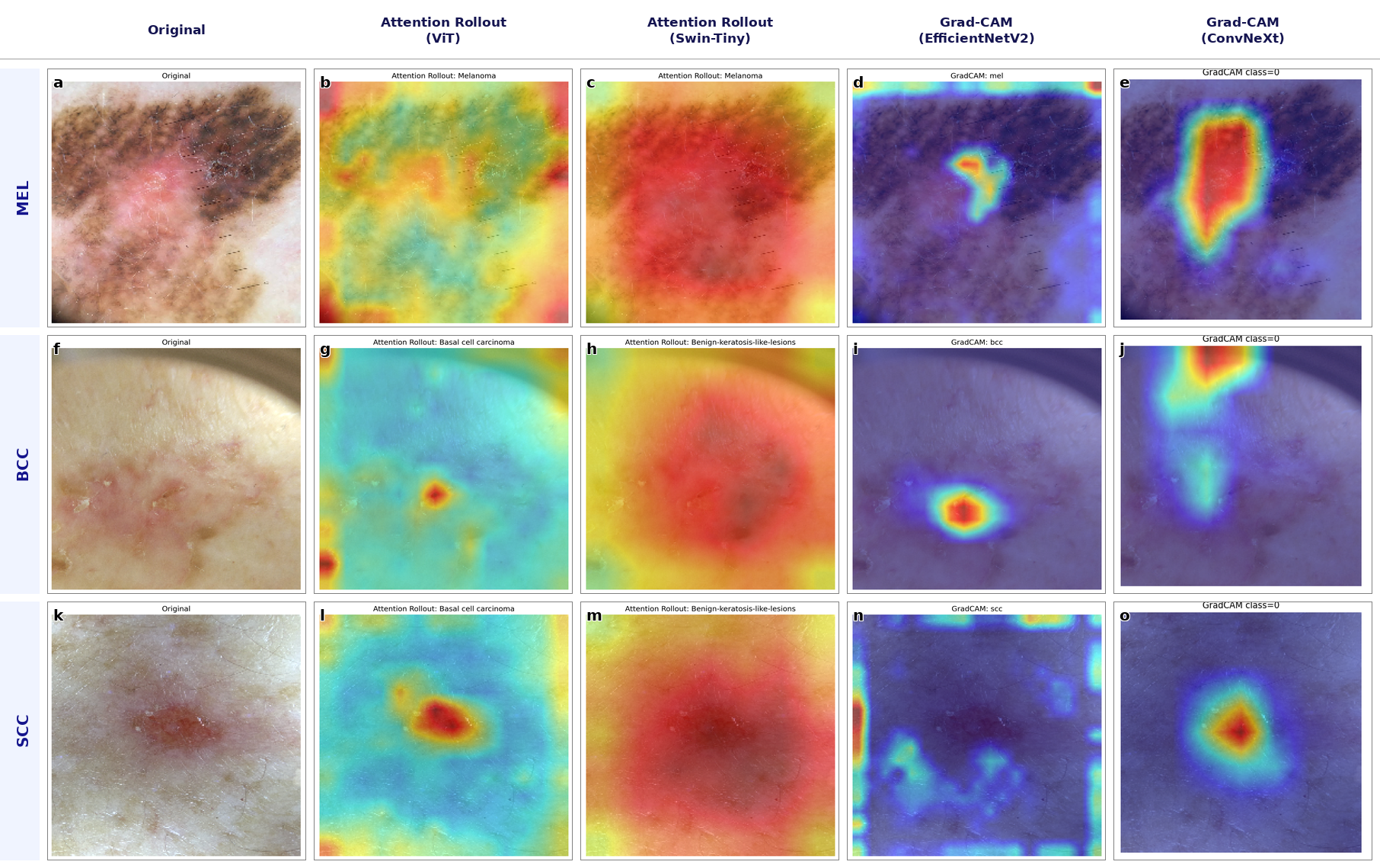}
\caption{Attention map examples for three nosological groups (rows: MEL\,---\,melanoma; BCC\,---\,basal cell carcinoma; SCC\,---\,squamous cell carcinoma). Columns: 1\,---\,original dermoscopic image; 2\,---\,Attention Rollout (ViT, Melanoma-Classification model); 3\,---\,Attention Rollout (Swin-T); 4\,---\,Grad-CAM (EfficientNetV2); 5\,---\,Grad-CAM (ConvNeXt). Warm colours (red, yellow) indicate regions of highest model activation. Panels: a--e\,---\,MEL; f--j\,---\,BCC; k--o\,---\,SCC.}\label{fig:attn}
\end{figure}

Figure~\ref{fig:attn} shows representative attention maps for three nosological groups from the validation sample compared across five visualisation methods. Transformer architectures (ViT, Swin) produce a~diffuse activation field covering the entire lesion structure, whereas convolutional models (EfficientNetV2, ConvNeXt) localise individual morphological elements, consistent with the IoU values obtained.

\subsection{Clinical Validation}\label{sec:apr}

\subsubsection{Study Design and Participants}

The clinical validation was conducted within a~series of ``Melanoma Day'' preventive screening sessions (\textnumero6\,--\,9) at Beauty Clinic (Orel). Study design: prospective case series with independent parallel expert assessment. The expert opinion was formed before the clinician gained access to the automatic classification result (blinding to system output). The panel of clinical experts comprised: Seregin~S.\,S., Cand.~Sci. (Med.), Chief Oncologist of the Orel Region\,---\,primary reference expert; Kozachok~E.\,S., dermatologist\,---\,second expert; and the clinic's general practitioner (GP)\,---\,third assessment level, used to measure the gain in diagnostic accuracy. The GP rendered an opinion twice on the same cases within each session: first independently without access to the system result (standard consultation protocol); second with access to the classification result and attention heatmap, after completing the initial examination of all session patients.

\subsubsection{Examination Protocol}

The patient participation protocol comprised the following steps: initial visual skin inspection with identification of the dermoscopy target; dermoscopic image acquisition with an optical dermatoscope coupled to a~smartphone in accordance with the standard operating procedure~\cite{kozachok2026dataset} (polarised mode, object centring, image quality control); independent expert opinion formed without access to the system result; automatic classification by \emph{Melanoscope AI} with recording of $P$, the Stage-2 nosological class, and the attention-map IoU value; when the routing algorithm (Section~\ref{sec:routing}) placed the patient in the red zone\,---\,referral for biopsy and histological examination with the result recorded in the referral registry. Inclusion criteria: age $\geq 18$~years, presence of a~skin lesion accessible for dermoscopy, informed consent to participate. Images with \texttt{image\_quality\_score}~$\leq 2$ and patients who had received treatment for the evaluated lesion less than 3~months before the session were excluded.

\subsubsection{Reference Standard}\label{sec:refstd}

The validation design employed differential verification: cases classified as malignant by the system or the expert (red-zone placement or clinical expert suspicion) were verified against histological examination of biopsy material as the reference standard (diagnostic gold standard); cases classified as benign by both methods were verified against the blinded independent clinical and dermoscopic assessment of two experts. This design corresponds to partial differential verification (\textit{partial verification bias}), accepted in screening studies where biopsy of all cases is not feasible. This constitutes a~study limitation (Section~\ref{sec:lim}) and may lead to systematic overestimation of specificity, as benign cases potentially misclassified as negative by the system are not histologically verified.

\subsubsection{Sample Characteristics}

The analytical sample comprised data from 4~sessions: 176~patients and 176~primary dermoscopic images (one image of the main evaluated lesion per patient). The session distribution is presented in Table~\ref{tab:sessions}.

\begin{table}\footnotesize
\caption{Clinical validation sessions}\label{tab:sessions}
\centering
\setlength{\tabcolsep}{3pt}
\renewcommand{\arraystretch}{1.2}
\begin{tabular}{|C{12mm}|C{22mm}|P{35mm}|C{20mm}|C{18mm}|}
\hline
Session & Date & Site & Patients & Images\\
\hline
\textnumero6 & 07.06.2025 & Beauty Clinic, Orel & 40 & 40\\
\thinhline
\textnumero7 & 17.10.2025 & Beauty Clinic, Orel & 43 & 43\\
\thinhline
\textnumero8 & 26.12.2025 & Beauty Clinic, Orel & 50 & 50\\
\thinhline
\textnumero9 & 24.04.2026 & Beauty Clinic, Orel & 43 & 43\\
\hline
\textbf{Total} & \textbf{June 2025\,--\,April 2026} & & \textbf{176} & \textbf{176}\\
\hline
\end{tabular}
\end{table}

Demographic characteristics of the validation sample: patient age range 19 to 84~years, median 42~years; 113~women (64.2\,\%) and 63~men (35.8\,\%); predominant Fitzpatrick skin phototypes\,---\,II (49.4\,\%) and III (37.5\,\%), consistent with the typical population of central Russia.

\subsection{Ethical Aspects}\label{sec:ethics}

The study was conducted in accordance with the Declaration of Helsinki. All patients provided informed consent to participate in the validation and to the processing of anonymised clinical data and dermoscopic images. The study was observational in character and did not alter the standard patient care pathway. System outputs were used as an auxiliary tool and did not replace the clinician's decision. Referral for biopsy was decided by the clinician on clinical grounds.

\section{Clinical Validation Results}\label{sec:results}

\subsection{Diagnostic Accuracy}

\begin{figure}
\centering
\includegraphics[width=0.85\linewidth]{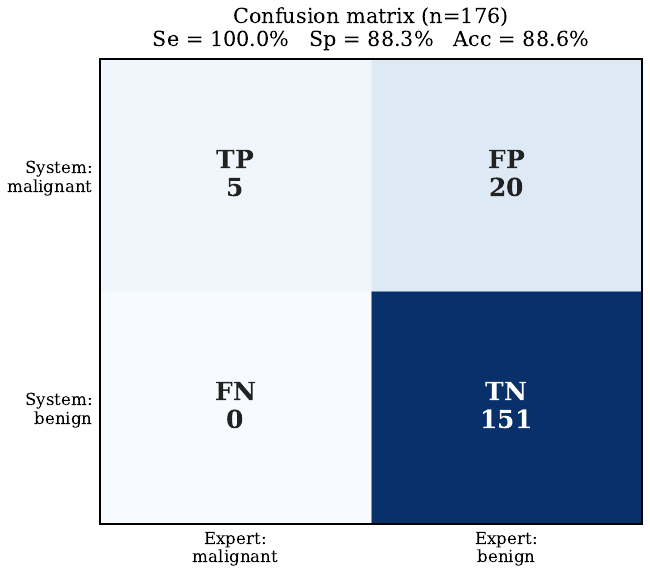}
\caption{Confusion matrix of the automatic classification ($n=176$). No false negatives were observed (FN$\,=\,$0).}\label{fig:cm}
\end{figure}

Agreement between the automatic classification and the independent expert opinion reached 88.6\,\% (156 agreements out of 176~images). The confusion matrix for the binary ``malignant / benign'' problem is presented in Table~\ref{tab:cm} and graphically in Figure~\ref{fig:cm}.

\begin{table}\footnotesize
\caption{Confusion matrix of the automatic classification ($n=176$)}\label{tab:cm}
\centering
\setlength{\tabcolsep}{3pt}
\renewcommand{\arraystretch}{1.2}
\begin{tabular}{|P{35mm}|C{20mm}|C{25mm}|C{15mm}|}
\hline
 & Expert: malignant & Expert: benign & Total\\
\hline
\textbf{System: malignant} & TP$\,=\,$5 & FP$\,=\,$20 & 25\\
\thinhline
\textbf{System: benign} & FN$\,=\,$0 & TN$\,=\,$151 & 151\\
\hline
\textbf{Total} & 5 & 171 & 176\\
\hline
\end{tabular}
\end{table}

Derived binary classification metrics: sensitivity\,---\,100.0\,\%; specificity\,---\,88.3\,\%; positive predictive value (PPV)\,---\,20.0\,\%; negative predictive value (NPV)\,---\,100.0\,\%; overall accuracy\,---\,88.6\,\%.

The low PPV (20.0\,\%) is expected for a~screening population with a~malignancy prevalence of 2.8\,\% (5/176): according to Bayes' theorem, at such a~low baseline risk even a~system with high sensitivity and 88\,\% specificity produces a~substantial proportion of false positives among patients referred to the red zone. This does not reduce the system's diagnostic value in a~screening context, but it means that on average 4 out of 5 red-zone patients have benign lesions. This must be explained to staff and patients, and taken into account when assessing the load on biopsy services.

Analysis of false-positive cases (FP$\,=\,$20): pigmented seborrhoeic keratoses with a~pseudo-reticular pattern ($n=9$), dysplastic naevi with marked atypia of the pigment network ($n=6$), and haemangiomas with a~dark pigmented crust ($n=5$). All false-positive cases fall into the red routing zone, ensuring their referral to a~specialist and thereby avoiding missed oncologically significant lesions\,---\,resulting only in excess dermatologist consultations. No false-negative cases were recorded in the validation sample (FN$\,=\,$0), consistent with the system design optimised for maximum sensitivity.

\subsection{Detected Lesions and Clinical Outcomes}

A~summary distribution of detected lesions and pathological conditions by validation session is presented in Table~\ref{tab:findings}.

\begin{table}\scriptsize
\caption{Detected lesions and pathological conditions by validation session}\label{tab:findings}
\centering
\setlength{\tabcolsep}{2pt}
\renewcommand{\arraystretch}{1.2}
\begin{tabular}{|P{52mm}|C{12mm}|C{12mm}|C{12mm}|C{12mm}|C{10mm}|}
\hline
Diagnosis\,/\,condition & \textnumero6 & \textnumero7 & \textnumero8 & \textnumero9 & Total\\
\hline
Melanoma (histol.\ confirmed) & \,---\, & \,---\, & 2 & 1 & \textbf{3}\\
\thinhline
Basal cell carcinoma (histol.\ confirmed) & \,---\, & 1 & \,---\, & 1 & \textbf{2}\\
\thinhline
Dysplastic naevus & 4 & 2 & \,---\, & \,---\, & \textbf{6}\\
\thinhline
\textbf{Oncol.\,/\,premalignant skin lesions} & \textbf{4} & \textbf{3} & \textbf{2} & \textbf{2} & \textbf{11}\\
\thinhline
Microcirculation disorders (lower limbs) & \,---\, & \,---\, & \,---\, & 2 & 2\\
\thinhline
Maxillary sinus pathology & \,---\, & \,---\, & \,---\, & 2 & 2\\
\thinhline
Cardiac arrhythmia & \,---\, & \,---\, & \,---\, & 1 & 1\\
\thinhline
\textbf{Systemic conditions} & \,---\, & \,---\, & \,---\, & \textbf{5} & \textbf{5}\\
\hline
\end{tabular}
\end{table}

The detection rate of oncological and premalignant skin lesions was $11/176 = 6.3\,\%$ of all participants examined in the validation.

\textbf{Melanoma.} In session~\textnumero8 the system assigned two patients to the red routing zone with $P\geq 0.50$ and nosological class~MEL at Stage~2; both patients were referred for biopsy and histological examination, and the diagnosis of melanoma was confirmed in both cases. In session~\textnumero9 one further melanoma case was detected ($P=0.71$, class~MEL), also histologically verified. All 3~melanoma cases fell into the red zone at Stage~1; none was missed by the system (melanoma sensitivity in the validation $=100\,\%$ at $n=3$; exact 95\,\% CI by Clopper--Pearson: 29.2\,--\,100.0\,\%).

\textbf{Basal cell carcinoma.} Two BCC cases were detected (one each in sessions~\textnumero7 and~\textnumero9), both assigned by the system to the red zone with Stage-2 class BCC and confirmed histologically (BCC sensitivity in the validation $=100\,\%$ at $n=2$; exact 95\,\% CI by Clopper--Pearson: 15.8\,--\,100.0\,\%).

\textbf{Dysplastic naevi.} Six dysplastic naevi with atypical dermoscopic features were detected (4 in session~\textnumero6 and 2 in session~\textnumero7). All were enrolled in a~dynamic follow-up programme with repeat dermoscopy at 6~months; none has shown signs of malignant transformation at the time of this publication.

\textbf{Session~\textnumero9\,---\,extended screening.} In the fourth session, five systemic pathological conditions were identified during an additional clinical examination (see lower part of Table~\ref{tab:findings}). These conditions are unrelated to the output of the automatic skin-lesion classifier but demonstrate the potential of the ``Melanoma Day'' format as a~platform for comprehensive primary screening under limited access to specialist care, and indicate prospects for expanding the system towards multimodal screening\,---\,integrating dermatological diagnosis with primary detection of systemic conditions.

\subsection{Effect on General Practitioner Diagnostic Accuracy}

Agreement between the GP's assessment and the assessment by Seregin~S.\,S. without access to the system results was 71.0\,\% (125 of 176); with joint use of the automatic classification result and the attention map, agreement rose to 82.4\,\% (145 of 176). The statistical significance of the difference was confirmed by McNemar's test ($p = 0.003$). The most pronounced gain was observed for cases with $P\in[0.15;\,0.50]$ (yellow routing zone), where the interpretable attention map allowed the clinician to justifiably reorient the diagnostic decision and shift from a~false-positive to a~more accurate assessment.

\section{Patient Routing Algorithm}\label{sec:routing}

\subsection{Rationale for the Three-Zone System}

The output of cascade Stage~1\,---\,the continuous malignancy probability $P$\,---\,cannot directly determine the clinical action without a~formal interpretation scheme. Standard clinical practice involves dividing a~continuous risk scale into discrete zones, each associated with a~standard staff action algorithm~\cite{argenziano2003}.

The threshold values $P=0.15$ and $P=0.50$ were derived from ROC analysis of the first cascade stage (sensitivity $\geq 0.95$ at $P=0.15$; Youden's optimal point at $P=0.50$) and confirmed by the clinical validation results: in the green zone ($P<0.15$) no false-negative cases were recorded in any of the four sessions; in the red zone ($P\geq 0.50$) all confirmed malignant lesions were detected ($5/5$).

\subsection{Algorithm Description}

\begin{figure}
\centering
\includegraphics[width=\linewidth]{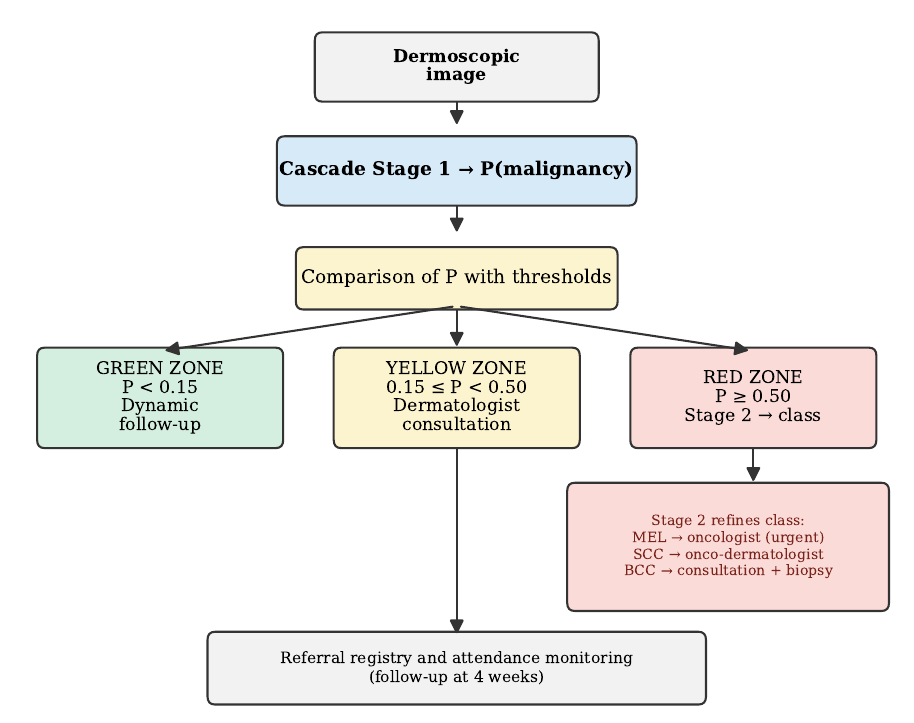}
\caption{Flowchart of the three-zone patient routing algorithm. $P(\text{malignancy})$ is computed; in the red zone the second stage refines the nosological class and defines the urgency of referral.}\label{fig:routing}
\end{figure}

The routing algorithm implements a~three-zone scheme (Figure~\ref{fig:routing}): \textbf{green zone} ($P<0.15$)\,---\,recording in the medical record, patient information, repeat examination in 6\,--\,12~months; \textbf{yellow zone} ($0.15\leq P<0.50$)\,---\,referral to a~dermatologist, repeat dermoscopy, biopsy if the patient falls into this zone again; \textbf{red zone} ($P\geq 0.50$)\,---\,urgent referral to an oncologist or onco-dermatologist, priority appointment, biopsy or excision.

For the red zone ($P\geq 0.50$), Stage~2 of the cascade refines the nosological class and determines the urgency of referral: class MEL (melanoma)\,---\,urgent referral to an oncologist (recommended within 3 working days); class SCC (squamous cell carcinoma)\,---\,referral to an onco-dermatologist; class BCC (basal cell carcinoma)\,---\,scheduled dermatologist consultation followed by biopsy.

\subsection{Application of the Algorithm in the Validation}

\begin{figure}
\centering
\includegraphics[width=0.85\linewidth]{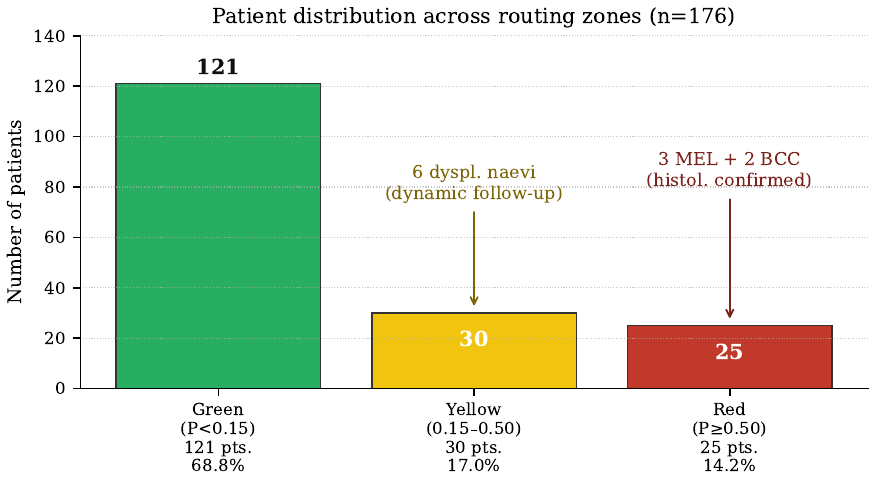}
\caption{Distribution of 176 validation-sample patients across the routing zones. The red zone contains 25 patients, including all 3 histologically confirmed melanomas and 2 BCCs.}\label{fig:zones}
\end{figure}

Based on the validation results, 176~patients were distributed across risk zones as follows (Figure~\ref{fig:zones}): green zone\,---\,121~patients (68.8\,\%), yellow\,---\,30~patients (17.0\,\%), red\,---\,25~patients (14.2\,\%). All 3~confirmed melanomas and 2~BCCs were assigned to the red zone at Stage~1 of the cascade. From the yellow zone, 6~patients with dysplastic naevi were referred for dynamic follow-up; none was confirmed as malignant at subsequent examination.

\section{Discussion}\label{sec:discussion}

\subsection{Comparison with Existing Clinical Decision Support Systems}\label{sec:compare}

The comparative analysis was performed against criteria relevant to assessing CDSS readiness for clinical use in Russian medical practice (Table~\ref{tab:compare}).

\begin{table}\scriptsize
\caption{Comparison of CDSSs for skin-lesion diagnosis}\label{tab:compare}
\centering
\setlength{\tabcolsep}{2pt}
\renewcommand{\arraystretch}{1.2}
\begin{tabular}{|P{50mm}|C{28mm}|C{20mm}|C{26mm}|C{20mm}|}
\hline
Criterion & \emph{Melanoscope~AI} (current) & SkinVision & Google Derm.\ Assist & Botkin.AI\\
\hline
Attention maps & $+$ rollout\,+\,Grad-CAM & $-$ & $-$ & $-$\\
\thinhline
Quantitative interpretability assessment (IoU) & $+$ & $-$ & $-$ & $-$\\
\thinhline
Standardised routing algorithm & $+$ 3 zones & $-$ & $-$ & $-$\\
\thinhline
Independent validation in Russia & $+$ 4 sessions 2025\,--\,2026 & $-$ & $-$ & Limited\\
\thinhline
Fitzpatrick phototypes I\,--\,IV (Russia) & $+$ & $-$ & $-$ & $-$\\
\thinhline
Mobile dermoscopy & $+$ & Partial & $-$ & $-$\\
\thinhline
Software registration in Russia & $+$ ISP~RAS & $-$ & $-$ & $+$\\
\thinhline
Melanoma sensitivity (valid.) & 100\,\% ($n{=}3$) & $\sim$91\,\%~\cite{udrea2020} & $\sim$83\,\%~\cite{liu2020} & N/A\\
\thinhline
Specificity (valid.) & 88.3\,\% & $\sim$79\,\%~\cite{udrea2020} & $\sim$76\,\%~\cite{liu2020} & N/A\\
\hline
\end{tabular}
\end{table}

The key advantage of the presented edition is the integrated chain ``cascade classification\,---\,attention map with IoU assessment\,---\,routing algorithm with justified thresholds'', no component of which is provided in full by the reviewed commercial counterparts. The 100\,\% melanoma sensitivity in the validation was achieved on a~small sample ($n=3$) and represents a~preliminary estimate; obtaining statistically reliable estimates requires expansion of the sample within a~multicentre study (Section~\ref{sec:lim}).

\subsection{Integration Recommendations}\label{sec:integration}

Recommendations were formulated based on the results of 4~validation sessions and are addressed to three types of healthcare facilities with varying resource levels.

\subsubsection{Dermatologist's Outpatient Office}

The CDSS is integrated into the standard consultation as an auxiliary tool: the specialist performs mobile dermoscopy, receives the classification result and attention map, and makes the clinical decision incorporating both sources. Technical requirements are minimal: a~smartphone, an optical dermatoscope, and internet access (or a~local network for offline deployment). Recommended performance benchmark: image analysis time not exceeding 15~seconds. Image requirements: resolution $\geq 1024\times 1024$ pixels, polarised mode, \texttt{image\_quality\_score}~$\geq 3$ on the standard operating procedure (SOP) scale~\cite{kozachok2026dataset}.

\subsubsection{Screening Programme (``Melanoma Day'')}

The validation session format optimally matches the screening use case. Recommended organisational parameters: staffing\,---\,1~oncologist or dermatologist (reference expert role) and 1\,--\,2~specialists with primary training in system use; throughput\,---\,up to 50~patients per session at a~rate of 5\,--\,7~minutes per patient; routing\,---\,direct issuance of consultation or biopsy referrals for red-zone patients at the session itself (eliminates patient loss through self-referral); maintenance of a~referral registry for red- and yellow-zone patients with attendance monitoring (control point\,---\,4~weeks).

\subsubsection{Resource-Limited Clinic (General Practitioner)}

Use of the system in facilities without a~staff dermatologist is most critical for expanding early-diagnosis coverage. In this context the routing algorithm functions as a~standardised staff action protocol: green zone\,---\,patient information, self-monitoring, recording in the medical record; yellow zone\,---\,telemedicine dermatologist consultation or scheduled specialist visit; red zone\,---\,priority referral to the regional dermatovenerological dispensary or oncology centre. A~specific recommendation for facilities of this type is mandatory pre-use staff training (2\,--\,4~hours based on a~standardised checklist) and periodic remote consultation of complex cases by a~regional-centre dermatologist.

\subsubsection{Phased Implementation Model}

Regardless of facility type, a~phased implementation model is recommended: preparatory stage (2\,--\,4~weeks)\,---\,staff training, equipment configuration, testing on a~control sample of 10\,--\,20~images with known diagnoses; pilot stage (1~month)\,---\,parallel operation of the system and the clinician with independent recording of both assessments, discrepancy analysis at the end of the month; full-scale stage\,---\,integration into the permanent clinical flow, quarterly accuracy audit (target agreement rate with expert assessment $\geq 85\,\%$).

\subsection{Limitations}\label{sec:lim}

The primary limitation of the presented validation is sample size: 176~patients and 176~primary dermoscopic images are sufficient to demonstrate system operability and provide an initial diagnostic accuracy estimate, but insufficient to obtain statistically robust estimates of sensitivity and specificity for individual rare nosological classes. Sensitivity values of 100\,\% for melanoma ($n=3$; 95\,\% CI: 29.2\,--\,100.0\,\%) and BCC ($n=2$; 95\,\% CI: 15.8\,--\,100.0\,\%) are preliminary and require confirmation in an expanded sample. The small number of malignant cases is inherent to the screening design: the sample was drawn from voluntary participants in a~preventive screening session rather than from patients with clinical indications, creating a~selection bias towards benign lesions (prevalence 2.8\,\%) and limiting generalisability to clinical populations with higher prior malignancy probability.

The second limitation is the partial verification design (Section~\ref{sec:refstd}): histological verification was obtained only for red-zone and clinically suspicious cases; benign cases were confirmed by expert clinical and dermoscopic assessment. This design is standard in screening studies, but creates a~risk of verification bias: true false negatives (malignant lesions not placed in the red zone) may remain undetected, potentially underestimating FN and overestimating specificity.

The third limitation is the single-centre validation design (Beauty Clinic, Orel); generalisability to other clinical settings, other dermoscopic equipment types, and other demographic groups requires multicentre verification.

The fourth limitation concerns the IoU metric: the reference standard consists of bounding rectangles rather than precise pixel-level masks of dermoscopic structures, which underestimates the absolute metric values and inflates variance; however, the architecture preference ordering remains stable.

The fifth limitation is the within-subject design for measuring GP accuracy gain: both measurements (before and after provision of the system result) were made by the same specialist on the same cases within a~single session. This design is standard for evaluating the impact of a~tool on diagnostic decision-making, but it carries the potential influence of the initial assessment on the subsequent one. Separating the initial and repeat examinations in time (after all initial session consultations are completed) reduces the likelihood of carry-over from the initial judgement. For more rigorous assessment, future studies may employ a~crossover design with a~washout interval between measurements, or parallel groups of clinicians.

The sixth limitation is a~structural conflict of interest: the first author is simultaneously the developer of the \emph{Melanoscope~AI} system, the second dermatologist expert in the validation, and the sole analyst responsible for data collection and analysis. Independent expert assessment of dermoscopic images and patient referral for biopsy were performed by Seregin~S.\,S. as the primary reference expert; histological verification of all detected malignant lesions was obtained independently. Nevertheless, the absence of an external data audit is a~limitation inherent in single-centre observational studies of this stage and should be addressed in an expanded multicentre study.

Future directions: expansion of the sample within a~multicentre study involving several regional facilities (target $\geq 500$ patients); adoption of a~crossover design for the GP accuracy-gain measurement with a~washout interval; quantitative assessment of the effect of the accuracy gain on clinical outcomes (stage of detected malignancy, time to treatment initiation); transition to pixel-level annotation of dermoscopic structures to improve IoU precision.

\section{Conclusion}\label{sec:conclusion}

The present work reports: a~new edition of the \emph{Melanoscope~AI} CDSS~\cite{kozachok2026reg} implementing two-stage cascade classification with a~mobile image-acquisition application and an attention-map visualisation module; a~quantitative interpretability assessment method for cascade models based on the IoU metric between high-activation maps and expert annotations of dermoscopic structures; results of a~prospective clinical validation with independent expert assessment in 4~``Melanoma Day'' preventive screening sessions (June~2025\,--\,April~2026, Beauty Clinic, Orel; 176~patients, 176~primary images); and a~three-zone patient routing algorithm with clinically justified thresholds $P<0.15$\,/\,$0.15$\,--\,$0.50$\,/\,$\geq 0.50$.

\textbf{Main results:}
\begin{enumerate}
\item Agreement between automatic classification and expert assessment\,---\,88.6\,\%; no false negatives were observed among 5 malignant lesions (5 of 5; 95\,\% CI by Clopper--Pearson: 47.8\,--\,100.0\,\%), specificity\,---\,88.3\,\%. All 3~melanoma cases and 2~BCC cases detected by the system were histologically verified.
\item ViT-B/16 with attention rollout shows the highest agreement of attention maps with expert patterns (mean IoU$\,=\,$0.69, exceeding convolutional architectures by 0.16\,--\,0.18).
\item Distribution of patients across routing zones (green 68.8\,\%, yellow 17.0\,\%, red 14.2\,\%) matches the expected specialist-load structure and is feasible within a~standard ``Melanoma Day'' screening session.
\item GP agreement with the expert assessment increases from 71.0\,\% to 82.4\,\% with access to the system output (McNemar's test, $p=0.003$).
\end{enumerate}

\textbf{Scientific novelty.} For the first time in Russian practice, a~quantitative method for assessing the agreement of attention maps of cascade deep learning models with expert dermoscopic structures has been proposed and applied in the context of independent clinical validation; a~three-zone patient routing algorithm with clinically justified malignancy probability thresholds has been developed\,---\,an original methodological contribution with no analogues in published domestic CDSSs for dermoscopy. The developed edition of \emph{Melanoscope AI} differs from prior publications~\cite{kozachok2025dataset,kozachok2025screening,kozachok2025sppvr1,kozachok2025sppvr2} in composition (mobile application, interpretability module), functionality (quantitative attention-map assessment, routing), and evidence base (4~validation sessions, histological verification of all malignant lesions).

\textbf{Practical significance.} Three operational integration scenarios are formulated (dermatologist's outpatient office, ``Melanoma Day'' screening programme, resource-limited clinic without a~staff dermatologist) together with a~phased implementation model based on the experience of 4~validation sessions. The proposed routing algorithm with specific probability thresholds ensures reproducibility of staff clinical actions and can be incorporated into the workflow of all three types of facilities without additional equipment requirements.

\section*{Authors' Contributions}\label{sec:contrib}

Kozachok~E.\,S.\,---\,study concept and design, development of the \emph{Melanoscope~AI} system and the IoU interpretability assessment method, data collection and analysis, manuscript writing and editing.

Seregin~S.\,S.\,---\,clinical validation, independent expert assessment of dermoscopic images, patient referral for biopsy, histological result verification.

\section*{Author Information}\label{sec:authors}

\noindent\textbf{Elena S. Kozachok}\,---\,specialist, Ivannikov Institute for System Programming of the Russian Academy of Sciences (ISP~RAS). ORCID:~0009-0007-9432-1663. E-mail: e.kozachok@ispras.ru

\noindent\textbf{Sergey S. Seregin}\,---\,Candidate of Sciences (Medicine), oncologist, Orel Regional Oncology Dispensary. ORCID:~0000-0002-7248-402X. E-mail: serega\_s2004@mail.ru

\section*{Acknowledgments}\label{sec:ack}

The authors thank the staff of Beauty Clinic (Orel) for organisational support of the ``Melanoma Day'' preventive screening sessions. The work was carried out within the research plan of the Ivannikov Institute for System Programming of the Russian Academy of Sciences.

\section*{Conflict of Interests}\label{sec:coi}

The authors declare no financial conflict of interests related to the publication of this article. The authors acknowledge a~structural conflict of interest: Kozachok~E.\,S. is concurrently the developer of the \emph{Melanoscope~AI} system, the second dermatologist expert in the clinical validation, and the sole analyst responsible for data collection and analysis. Independent primary expert assessment of dermoscopic images, patient referral for biopsy, and histological verification were performed independently by Seregin~S.\,S.

\end{document}